\documentclass[runningheads]{llncs}
\usepackage{graphicx}
\usepackage{amsmath}
\usepackage{amssymb}
\usepackage{diagbox}
\usepackage{multirow}
\usepackage{bbding}
\begin{document}
\title{Multi-layer Sequence Labeling-based Joint Biomedical Event Extraction}
% \title{Multi-layer Sequence Labeling-based Joint Biomedical Event Extraction\thanks{Supported by organization x.}}
\author{
Gongchi Chen\inst{1}\orcidID{0009-0007-6037-7669} \and
Pengchao Wu\inst{1} \and
Jinghang Gu\inst{2} \and
\\ 
Longhua Qian \inst{1}\Envelope \and
Guodong Zhou \inst{1}
}

\authorrunning{G. Chen et al.}

\institute{
School of Computer Science and Technology, Soochow University, Suzhou, Jiangsu Province 215006, China \\
\email{\{20224227020, 20204227037\}@stu.suda.edu.cn, \{qianlonghua,gdzhou\}@suda.edu.cn}\\
\and
Department of Chinese and Bilingual Studies, The Hong Kong Polytechnic University, Hong Kong 999077, China\\
\email{gujinghangnlp@gmail.com}
}
\maketitle              % typeset the header of the contribution
\begin{abstract}
In recent years, biomedical event extraction has been dominated by complicated pipeline and joint methods, which need to be simplified. In addition, existing work has not effectively utilized trigger word information explicitly. Hence, we propose MLSL, a method based on multi-layer sequence labeling for joint biomedical event extraction. MLSL does not introduce prior knowledge and complex structures. Moreover, it explicitly incorporates the information of candidate trigger words into the sequence labeling to learn the interaction relationships between trigger words and argument roles. Based on this, MLSL can learn well with just a simple workflow. Extensive experimentation demonstrates the superiority of MLSL in terms of extraction performance compared to other state-of-the-art methods.

\end{abstract}

\section{Introduction}
\begin{figure}
\vspace{-0.5cm}
\centerline{\includegraphics[width=0.8\linewidth]{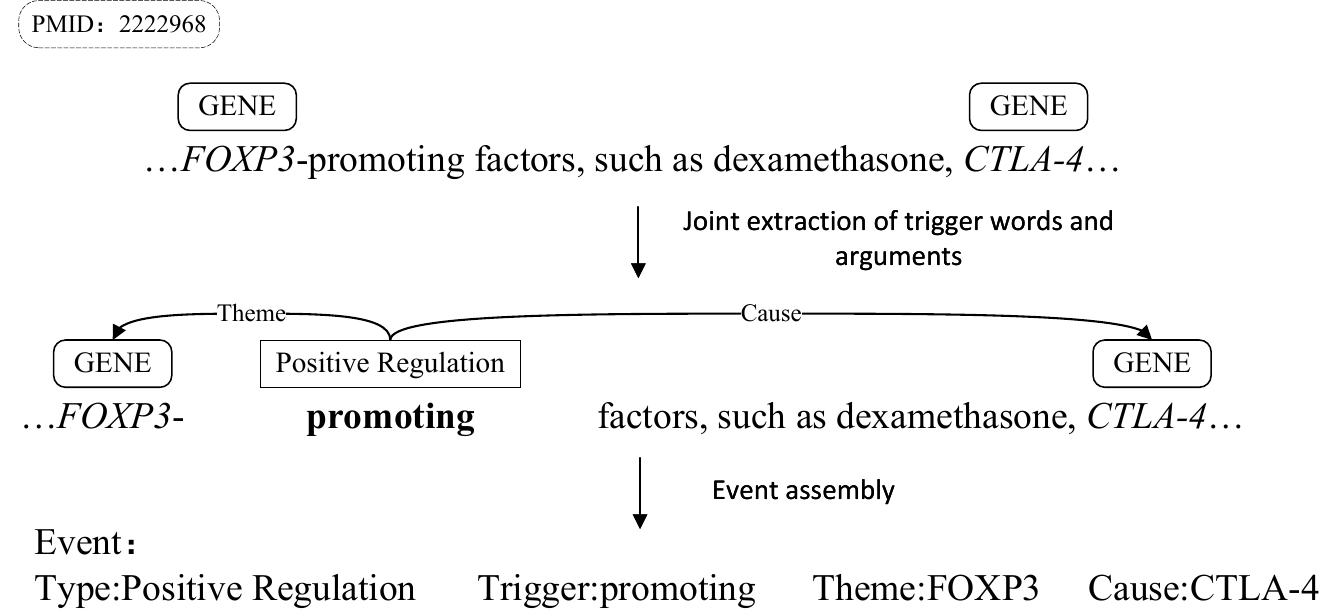}}
\caption{Overall workflow of biomedical event extraction.} \label{fig1}
\vspace{-0.5cm}
\end{figure}

Biomedical event extraction (i.e., BEE) is an essential task for extracting key information from the massive biomedical literature, receiving mounting attention in the NLP community in recent decades \cite{frisoni2021survey}. The overall workflow of BEE is shown in Fig.~\ref{fig1}. Given an input sentence (e.g., "... FOXP3 promoting factors, such as dexamethasone, CTLA-4..."), the entity mentions (e.g., "FOXP3" and "CTLA-4" with "GENE" type), BEE needs to recognize the trigger word (e.g., "promoting" with the "Positive Regulation" type) and the corresponding arguments (e.g., a theme argument "FOXP3" and an optional cause argument "CTLA-4") simultaneously. Afterwards, the final biomedical events (e.g., two "Positive Regulation" events \textless PoRe, promoting, cause, CTLA-4\textgreater and \textless PoRe, promoting, theme, FOXP3\textgreater) are obtained by a simple assembling rule: a regulation event must have a theme argument and an optional cause argument.

To address the task of BEE, many effective yet complicated pipeline and joint methods are proposed~\cite{frisoni2021survey}. In the pipeline manner, the results of a trigger word recognition model serve as input for an argument role recognition model, both of which are trained separately and do not affect each other~\cite{bjorne2018biomedical,li2019biomedical}. This manner is flexible and easy to understand, yet it ignores the inter-dependencies between subtasks and suffers from error propagation \cite{wang2022conditional}. Instead, the joint manner in BEE combines trigger word recognition with argument role recognition for training together to overcome the shortcomings of pipeline-based methods, by sharing parameters and information between the subtasks. Nevertheless, struggling to model the extraction of nested events, some researchers have to introduce complex structures (e.g., directed acyclic graph \cite{trieu2020deepeventmine}) or prior knowledge (e.g., dependency parsing \cite{zhao2021novel}) into their model architectures or workflows.

In a bid to simplify the workflow, we propose a method based on multi-layer sequence labeling for joint biomedical event extraction, namely MLSL. It's a data-driven method that does not introduce complex prior knowledge such as knowledge graphs or dependency parsing, while it only learns from the input data. Additionally, we designed MLSL in a pipeline paradigm so that it can first recognize trigger words using a trigger layer, and then explicitly merge the information of those candidate trigger words into the hidden representation. Empirical results show that the MLSL without complex structures outperforms other state-of-the-art methods on BEE. In addition, the information of trigger words can be very useful to assist the argument recognition in order to improve the overall performance of BEE.

\section{Methods}
\subsection{Overall workflow of MLSL}
The ultimate goal of joint learning of trigger recognition and argument recognition is to extract biomedical events in the form of a quadruple like \textless type, trigger word, argument role, argument\textgreater. There are usually two types of argument roles~\cite{frisoni2021survey}: one is the \textit{theme} role, which refers to arguments whose attributes are altered by the event, and the other is the \textit{cause} role, which refers to arguments that cause the event to occur. For simplification , we cast the task as a multi-task sequence labeling problem. Formally, given an input token sequence $X=\left\{x_1, x_2, \ldots, x_n \right\}$, MLSL is modeled as $X\mapsto f_{C/T}(Y_{C/T}|Y_{TR})$, where $Y_{TR}$ is derived from $f_{TR}: X\mapsto Y_{TR}$. Here, $f_{TR}$ is the trigger layer learning to recognize trigger words, $f_C$ is the cause layer learning to recognize cause arguments, and $f_T$ is the theme layer learning to recognize theme arguments. $Y_{TR}$, $Y_C$ and $Y_T$ are the trigger word labels, cause argument labels, and theme argument labels assigned to each token, respectively.

\begin{figure}
\vspace{-0.5cm}
\centerline{\includegraphics[width=0.9\linewidth]{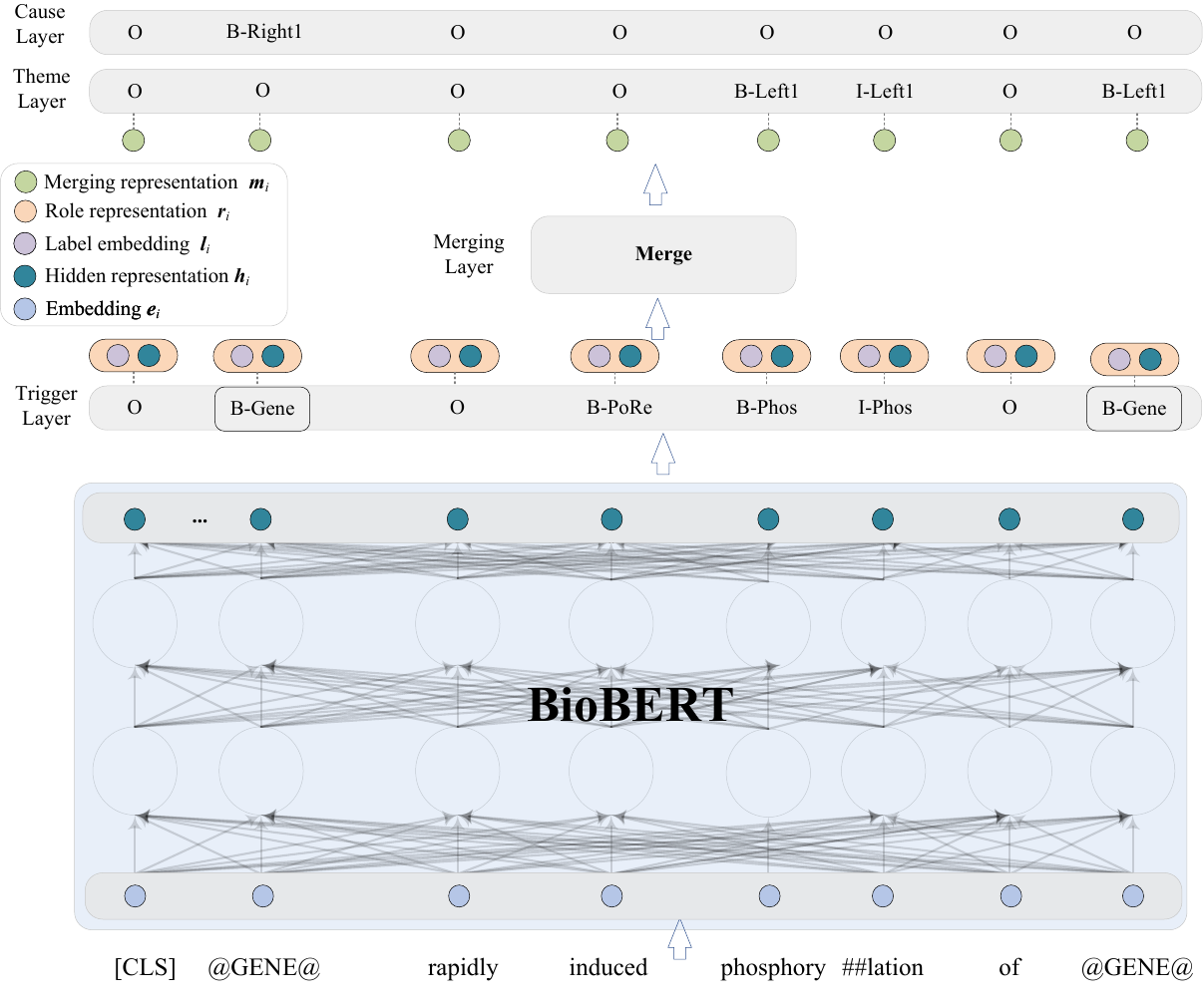}}
\caption{Multi-layer sequence labeling-based joint biomedical event extraction model} \label{fig2}
\vspace{-0.5cm}
\end{figure}

Following the above principle, we construct the overall workflow of MLSL as shown in Fig.~\ref{fig2}. Given an input sequence (e.g., "BMP-6 rapidly induced phosphorylation of Smad1/5/8") and corresponding entity mentions (e.g., "BMP-6" and "Smad1/5/8" with "GENE" type) as an example, we first tokenize the input sentence to get a token sequence, and replace entity mentions with special tokens (e.g., the token "@GENE@" for the "GENE" type) for better generalization~\cite{alt2018improving}. Next, we get the embeddings $E=\left\{e_1, e_2, \ldots, e_n\right\}$ of the input token sequence and utilize a BioBERT~\cite{lee2020biobert} encoder to get its hidden representation $H=\left\{h_1, h_2, \ldots, h_n\right\}$, where $e_i, h_i \in \mathbb{R}^{d}$. Note that \textit{d} is the hidden dimension size of BioBERT. Then, a trigger layer is used to recognize the labels of trigger words (e.g., "B-Phos" and "I-Phos" are predicted), while the labels of given entity mentions do not need to be recognized (e.g., "B-Gene" is given). Meanwhile, the labels of other tokens are set to "O". Afterwards, we can get the role representation $R=\left\{r_1, r_2, \ldots, r_n \right\}$ of the input token sequence, where $r_i=\operatorname{concat}(l_i, h_i)$. It's worth noting that $l_i$ is \textit{i}-th label embedding of the label assigned to the \textit{i}-th token (e.g., $l_1$ is the embedding of the "B-Gene" label, $l_3$ is the embedding of the "B-PoRe" label). Later, we explicitly employ the information of trigger words in the merging layer to get the merging representation $M=\left\{m_1, m_2, \ldots, m_n \right\}$, whose detailed processing is illustrated in Sec.~\ref{merging_layer}. After that, a theme layer and a cause layer recognize the theme arguments and the cause arguments respectively. Finally, we can assemble the identified trigger words, theme arguments, and cause arguments into event results based on rule-based event construction~\cite{wu2024pipelined}.

\subsection{Labeling schema}\label{labeling_schema}
Our proposed MLSL has a multi-layer sequence labeling decoder consisting of the trigger layer, the theme layer and the cause layer, each of which decodes (i.e., predicts) a specific type of labels using the BIO labeling schema to easily convert the biomedical event into the multi-layer sequence labeling representation.

\begin{figure}
\vspace{-0.5cm}
\includegraphics[width=\textwidth]{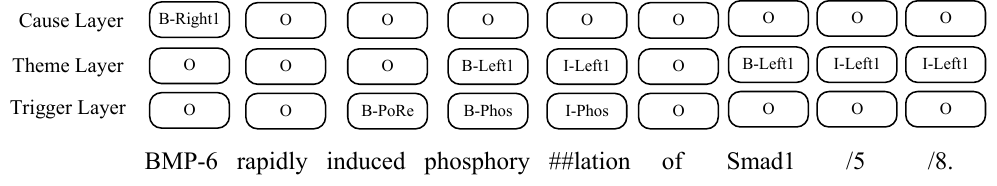}
\caption{An example of labeling schema} \label{fig3}
\vspace{-0.5cm}
\end{figure}

As illustrated in Fig.~\ref{fig3}, the BIO schema is sufficient to label different types of trigger words. For instance, the trigger word "phosphorylation" with the type of "phosphorylation" is tokenized into the tokens "phosphory" and "\#\#lation", each of which is assigned to the "B-Phos" label and the "I-Phos" label respectively. Similarly, the trigger word "induced" with the type of "Positive Regulation" is assigned to the "B-PoRe" label.

For argument labeling, we use the four labels of "Left1", "Left2", "Right1", and "Right2" to label the position of argument roles relative to a specific trigger word. In detail, "Left1", "Left2", "Right1", and "Right2" respectively indicate that the trigger word belonging to the argument is the 1st/2nd trigger word on its left/right. In other words, arguments that have more than 2 other trigger words away will be discarded. That's because those arguments only account for $\sim4$\%/$\sim3$\% in the GE11 train/dev set and $\sim3$\%/$\sim5$\% in the GE13 train/dev set according to our statistics. Furthermore, these arguments can be hardly recognized due to their long distance away from the trigger words in our preliminary experiment. Consequently, discarding these arguments will not have a major impact on the event extraction performance. Fig.~\ref{fig3} shows an example of the theme/cause arguments, where the token "@GENE@" next to the token "[CLS]" is assigned to the "B-Right1" label, indicating that "@GENE@" is the cause argument of the 1st right trigger word "induced". Likewise, the word "Smad1/5/8" which is tokenized into the token "Smad1", the token "/5" and the token "/8", is the theme argument of 1st left trigger word "phosphorylation". Therefore, "Smad1", "/5" and "/8" are respectively assigned to the "B-Left1" label, "I-Left1" label and "I-Left1" label.

\subsection{Multi-layer sequence labeling}\label{multi-layer_decoder}
As shown in Fig.~\ref{fig2}, we decode the arguments and trigger words using the multi-layer sequence labeling manner, including the trigger layer, the merging layer, the cause layer and the theme layer.

\subsubsection{Trigger layer}\label{trigger_layer}
The operation of the trigger layer is shown in the Eq. (\ref{trigger_prop}) and Eq. (\ref{trigger_pred}). The hidden representation $h_i \in \mathbb{R}^d$ of the \textit{i}-th token is fed into a full-connected layer and then a softmax operation, to obtain the probabilities $p_{tr}$ of trigger labels. After that, the trigger label with the largest probability $\hat{y}_{tr}$ is selected as the result of the \textit{i}-th token.
\begin{equation}\label{trigger_prop}
p_{tr} = \operatorname{softmax}(W_{tr}^{\top}h_i+b_{tr})
\vspace{-0.5cm}
\end{equation}

\begin{equation}\label{trigger_pred}
\hat{y}_{tr} = \operatorname{argmax}(p_{tr})
\end{equation}
where, $W_{tr}\in \mathbb{R}^{d\times \lvert Y_{TR}\rvert}$ and $b_{tr}\in \mathbb{R}^d$ are the weight and bias of the trigger layer respectively. $\lvert Y_{TR}\rvert$ is the label space of the trigger words.

\subsubsection{Merging layer}\label{merging_layer}
The merging layer is used to merge the information of the trigger word predicted by the trigger layer for the subsequent theme/cause layer. For each token $x_i$, we collectively refer the trigger words predicted not more than two on its left and those not more than two on its right in the trigger layer as its corresponding candidate trigger words. The set of the tokens in candidate trigger words is denoted as $C_i$. Additionally, we assign label embedding $l_i\in \mathbb{R}^{d}$ of the trigger word labels (e.g., "B-Phos" label, "O" label) or entity labels ("B-Gene" label) to each token. It is notable that the label embedding is randomly initialized for each trigger word label and entity label, and it will be updated during the training process. Formally, for each token $x_i$ in the merging layer, we first concatenate its label embedding $l_i$ and its hidden representation $h_i$ to get its role representation $r_i$:
\begin{equation}\label{role_rep}
    r_i = \operatorname{concat}(l_i,h_i)
\end{equation}
To explicitly employ such information, for each token $x_i$, we first obtain the representation of $C_i$ in three different ways, and then concatenate it with the hidden representation $h_i$ to obtain the merging representation $m_i$:
\begin{itemize}
    \item Average: As described in Eq. (\ref{aver}) and Eq. (\ref{merging_rep}), we concatenate the average of the role representation of $C_i$ to its hidden representation $h_i$, to get the merging representation $m_i$ of the token $x_i$.
    
    \item Attention: As illustrated in Eq. (\ref{att}) and Eq. (\ref{merging_rep}), for each token (e.g., $x_i$), we first calculate the attention weights between its role representation $r_i$ and the role representation of $C_i$. Then, we concatenate the weighted sum of the role representation of $C_i$ to the hidden representation $h_i$ to, get the merging representation $m_i$.
    
    \item Self-attention: Shown in Eq. (\ref{self_att}), for each token (e.g., $x_i$), we first calculate the self-attention~\cite{vaswani2017attention} between its role representation $r_i$ and the role representation of $C_i$ (i.e., $R_i^C \in \mathbb{R}^{d\times \lvert C_i \rvert}$) in the \textit{j}-th attention head. In Eq. (\ref{self_att}), $W_Q,W_K,W_V \in \mathbb{R}^{d_h\times d}$ are the query matrix, the key matrix and the value matrix respectively, and $d_h$ is the hidden size of an attention head. Next, we concatenate all results from the whole \textit{k} attention heads in Eq. (\ref{cat_multi_head}). Later in Eq. (\ref{merging_rep}), we obtain the merging representation $m_i$ by concatenating the self-attention weighted sum of the role representation of $C_i$ to the hidden representation $h_i$. 
\end{itemize}

{
\setlength{\abovedisplayskip}{1cm}
\setlength{\belowdisplayskip}{1cm}
\begin{equation}\label{aver}
    \widetilde{r}_{C_i} = \frac{1}{\lvert C_i \rvert}\sum_{j\in C_i}r_j
\end{equation}
}

\begin{equation}\label{att}
    \widetilde{r}_{C_i} = \sum_{j\neq i, j\in C_i}\frac{e^{r_i^\top r_j}}{\sum_{j\neq i, j\in C_i}e^{r_i^\top r_j}}r_j
\end{equation}

\begin{equation}\label{self_att}
    \hat{r}^j_{C_i} = (W_VR_i^C )\operatorname{softmax}(\frac{(W_{k}R_i^C)^\top(W_{Q}r_i)}{\sqrt{d_h}})
\vspace{-0.3cm}
\end{equation}

\begin{equation}\label{cat_multi_head}
    \widetilde{r}_{C_i} = \operatorname{concat}(\hat{r}^1_{C_i},\ldots, \hat{r}^k_{C_i})
\vspace{-0.5cm}
\end{equation}

\begin{equation}\label{merging_rep}
    m_i = \operatorname{concat}(h_i, \widetilde{r}_{C_i})
\vspace{-0.5cm}
\end{equation}

\subsubsection{Theme/Cause layer}\label{theme/cause_layer}
Based on the merging representation output from the merging layer, the theme layer and the cause layer predict theme arguments and cause arguments respectively. We first calculate the probabilities $p_{t/c}$ of theme/cause argument labels in (\ref{theme_cause_prop}). Then, we obtain the theme/cause argument labels by selecting the largest probability $\hat{y}_{t/c}$ as the result of the \textit{i}-th token.

\begin{equation}\label{theme_cause_prop}
p_{t/c} = \operatorname{softmax}(W_{t/c}^{\top}m_i+b_{t/c})
\vspace{-0.5cm}
\end{equation}

\begin{equation}\label{theme_cause_pred}
\hat{y}_{t/c} = \operatorname{argmax}(p_{t/c})
\end{equation}
where $W_{t}\in \mathbb{R}^{d\times \lvert Y_{T}\rvert}$ and $W_{c}\in \mathbb{R}^{d\times \lvert Y_{C}\rvert}$ are the weights of the theme layer and the  cause layer respectively. And $b_{t}, b_{c}\in \mathbb{R}^d$ are biases of the theme layer and the cause layer respectively.

\subsubsection{Multi-layer loss}
The trigger layer, the theme layer and the cause layer each perform a multi classification task separately. Therefore, we utilize cross-entropy to train each of them. The total loss of the MLSL is the sum of the losses of the three layers, i.e., $L=L_{tr}+L_{t}+L_{c}$, which is a multi-task learning loss.

\section{Experiments}
\subsection{Setup}

\subsubsection{Datasets}
We train and evaluate our MLSL on the BioNLP11-GE\footnote{https://2011.bionlp-st.org/bionlp-shared-task-2011}(i.e., GE11) ~\cite{kim2011overview} and BioNLP13-GE (i.e., GE13)\footnote{https://2013.bionlp-st.org/introduction} ~\cite{kim2013genia}, whose specific statistics are illustrated in Table~\ref{data_stas}. Both GE11 and GE13 provide entity mentions for each sentence. Hence, we should predict the correct events for each sentence given the entity mentions.

\begin{table}
\vspace{-0.3cm}
\centering
\caption{The statistics of the GE11 and GE13 event dataset}
\label{data_stas}
\setlength{\tabcolsep}{1.5mm}{
\begin{tabular}{l|rrr|rrr}
\hline
\multirow{2}{*}{\textbf{Item}} & \multicolumn{3}{c|}{\textbf{GE11}} & \multicolumn{3}{c}{\textbf{GE13}} \\
\cline{2-7}
& \textbf{Train} & \textbf{Dev} & \textbf{Test} & \textbf{Train} & \textbf{Dev} & \textbf{Test} \\
\hline
Sentences & 8664 & 2888 & 3363 & 2438 & 2727 & 3574 \\
Events & 10310 & 3250 & 4487 & 2817 & 3199 & 3348 \\
Entities & 11652 & 4690 & 5301 & 3692 & 4452 & 4686 \\
\hline
\end{tabular}
}
\vspace{-1cm}
\end{table}

\subsubsection{Evaluation}
The official BioNLP shared tasks provide online evaluation  tools\footnote{http://bionlp-st.dbcls.jp/GE/2011/eval-test/}. Usually, the results of the approximate span matching and approximate recursive matching are used as the final results of the test set. Approximate span is the predicted span that can be different from the gold span within a single token. Approximate recursive is a rule for nested events, under which even if a sub-event is partially correct, it is considered a correct event. Three standard metrics including precision (P), recall (R) and micro-averaging F1-score (F1) are used to evaluate performance. We train and validate the MLSL 5 times using different random seeds, and select the checkpoint with the best results on the dev set for subsequent evaluation, where the average scores of 5 runs with different random seeds are adopted as the final results.

\begin{table}
\vspace{-0.5cm}
\centering
\caption{Hyper-parameter values and search space}
\label{hyper_value}
\setlength{\tabcolsep}{1.5mm}{
\begin{tabular}{lrr}
\hline
\textbf{Hyper-pamameter} & Value & Space \\
\hline
Optimizer & Adam & \\
Encoder &biobert\_v1.1\_pubmed\footnote{https://huggingface.co/monologg/biobert\_v1.1\_pubmed} & \\
$\beta_1$,$\beta_2$ & 0.9, 0.99 & \\
Dropout & 0.3 & {0.1, 0.3, 0.5} \\
Layer dropout &0.1 & \\
Batch size &32 & {8, 16, 32}\\
Learning rate &1e-5 & {1e-3, 1e-4, 1e-5} \\
Epoch &20 &{10, 20} \\
\hline
\end{tabular}
}
\vspace{-1cm}
\end{table}

\subsubsection{Setting}
We implement our model using PyTorch\footnote{https://pytorch.org/} and transformers\footnote{https://huggingface.co/docs/transformers/index}. The Hyper-parameter values and search space are listed in Table~\ref{hyper_value}. 

\subsection{Comparison of different merging method}
Table~\ref{ablation_res} shows the micro F1 results of using different merging methods in different tasks of the GE11 and GE13 development sets. Here, the "None" method means that there is no merging layer to merge the information of the candidate trigger words. "Trg", "Arg" and "Eve" indicate the trigger extraction task, the argument extraction task and the event extraction task, respectively. "Avg", "Att" and "Self-att" are short for the \textit{average} merging method, the \textit{attention} merging method and the \textit{self-attention} merging method, respectively. It can be found that the \textit{self-attention} merging method achieves the best performance in all tasks, compared to the other merging methods. This may be due to the use of more trainable parameters (e.g., $W_Q$, $W_K$ and $W_V$) in this method, allowing the role representation of each token to better focus on and integrate information from its candidate trigger words. Accordingly, we choose the \textit{self-attention} merging layer for our MLSL to compare with other baselines.

\begin{table}
\vspace{-0.5cm}
\centering
\caption{The performance comparison of using different merging methods on the dev sets. \textit{Note}: the numbers in brackets are standard deviation of 5 runs.}
\label{ablation_res}
\resizebox{\textwidth}{!}{
\begin{tabular}{lccc|ccc}
\hline
\multirow{2}{*}{\textbf{Setting}} & \multicolumn{3}{c}{\textbf{GE11}} & \multicolumn{3}{c}{\textbf{GE13}} \\
\cline{2-7}
& \textbf{Trg}(\%) & \textbf{Arg}(\%) & \textbf{Eve}(\%) & \textbf{Trg}(\%) & Arg(\%) & \textbf{Eve}(\%) \\
\hline
None &76.64(±0.23) &73.86(±0.45) &58.24(±0.16) &77.66(±0.40) &68.12(±0.41) &55.00(±0.38) \\
 +Avg &76.93(±0.44) &74.15(±0.43) &58.91(±0.44) &77.86(±0.46) &68.20(±0.80) &55.80(±0.58) \\
 +Att &76.95(±0.40) &75.60(±0.32) &59.07(±0.33) &78.03(±0.58) &70.90(±0.35) &55.92(±0.64) \\
 +Self-att &\textbf{77.14}(±0.58) &\textbf{75.72}(±0.32) &\textbf{59.34}(±0.26) &\textbf{78.55}(±0.15) &\textbf{71.06}(±0.73) &\textbf{56.34}(±0.26)\\
\hline
\end{tabular}
}
\vspace{-1cm}
\end{table}

\subsection{Comparison with other systems}
In order to demonstrate the effectiveness of our proposed MLSL, we select several representative pipeline and joint methods for biomedical event extraction as baselines, which are listed below. 
\begin{itemize}
    \item TEES-CNN~\cite{bjorne2018biomedical}: A pipeline system for event extraction that operates by sequentially carrying out the extraction of entities, arguments, and events, with each component utilizing a CNN-driven sentence encoding framework.
    
    \item KBTL~\cite{li2019biomedical}: The KB-Tree LSTM model (short for KBTL) is a pipeline method that introduces an external knowledge base to the tree structured LSTM to enhance the semantic representation of words. 
    
    \item Wu et al.~\cite{wu2024pipelined}: A pipeline approach sequentially performing trigger identification, argument roles recognition and final event construction, which employ a n-ary relation extraction method to alleviate errors in event construction.  
    
    \item DeepEventMine~\cite{trieu2020deepeventmine}: A joint end-to-end method for nested event extraction, which is capable of extracting multiple, intersecting directed acyclic graph (DAG) structures directly from the raw text.
    
    \item CPJE~\cite{wang2022conditional}: A joint system for event extraction that uses the Graph Convolutional Neural Networks (i.e., GCN)~\cite{kipf2016semi} to model the dependency information.
    
    \item Zhao et al.~\cite{zhao2021improved}: An improved RL-based framework for multiple biomedical event extraction. It employs a self-supervised-based data augmentation method for biomedical entities and event triggers in raw texts. 
\end{itemize}

The comparison results on the test set are shown in Table~\ref{comparison}. It can be observed that our proposed MLSL outperforms other baselines in F1 score on both GE11 and GE13 thanks to the improvement of recall. Such a result shows the effectiveness of our MLSL. Different from other baselines, the MLSL does not use prior knowledge (e.g., knowledge base in KBTL), complex structures (e.g., graph in DeepEventMine or CPJE) or some sophisticated learning strategy(e.g., RL in Zhao et al. or n-ary relation extraction method in Wu et al.). It extracts triggers and arguments in a multi-layer sequence labeling schema, which can simplify the biomedical event extraction task. On the other hand, the MLSL incorporates candidate trigger words explicitly in the sequence labeling, which may enhance information interaction between the theme/cause arguments recognition task and the trigger word recognition task.

\begin{table}
\vspace{-0.5cm}
\centering
\caption{The performance comparison of different methods on the test sets}
\label{comparison}
\setlength{\tabcolsep}{1.5mm}{
\begin{tabular}{clccc|ccc}
\hline
\multirow{2}{*}{\bfseries {Manner}} &\multirow{2}{*}{\bfseries {Methods}} & \multicolumn{3}{c}{\bfseries {GE11}} & \multicolumn{3}{c}{\bfseries {GE13}} \\
\cline{3-8}
& & \bfseries{P}(\%) & \bfseries{R}(\%) & \bfseries{F1}(\%) & \bfseries{P}(\%) &\bfseries{R}(\%) & \bfseries{F1}(\%) \\
\hline
\multirow{3}{*}{pipeline} & TEES-CNN &69.45 &49.94 &58.10 &\textbf{65.78} &44.38 &53.00 \\
& KBTL &67.01 &52.14 &58.65 &62.01 &51.03 &55.99 \\
& Wu et al. &67.04 &\textbf{59.66} &63.14 &63.90 &55.50 &59.40 \\
\hline
\multirow{5}{*}{joint} &DeepEventMine &71.71 &56.20 &63.02 &60.98 &49.80 &54.83 \\
&CPJE &\textbf{72.62} &53.33 &61.50 &- &- &- \\
&Zhao et al. &- &- &- &64.21 &53.77 &58.53 \\
\cline{2-8}
&our MLSL &69.94 &59.50 &\textbf{64.30} &64.90 &\textbf{56.53} &\textbf{60.43} \\
\hline
\end{tabular}
}
\vspace{-1cm}
\end{table}

\subsection{Discussion and analysis}
\subsubsection{Empirical analysis}
For the purpose of further analyzing the recognition performance of the MLSL, we report its performance for recognizing different types of trigger words in Table~\ref{trigger_recognition}. There are 9 types of events in GE11, while GE13 adds an additional 4 types of events (i.e., "PrMo", "Ubiq", "Acet" and "Deac"). However, these 4 types of events are too sparse, accounting for less than 1\% in the train, dev, and test sets. Therefore, we will not list their recognition performance. In Table~\ref{trigger_recognition}, we can see that the MLSL cannot effectively identify the "Tran" event and the "Regu" event, resulting in a decrease in overall performance. For the "Tran" event, it accounts for a relatively small proportion compared to other simple events (e.g., "GeEx", "PrCa", "Phos" and "Loc"), so that the data-driven MLSL cannot effectively recognize the "Tran" type event. In terms of the "Regu" event, it is a nested event containing theme and cause (optional) argument. 

To analyze the performance of the "Regu" event, we record the performance of the MLSL for identifying different argument roles in Table~\ref{argument_recognition}. We can observe that the MLSL is able to better recognize the theme arguments than the cause arguments. This may be due to the lower proportion of cause arguments in the dataset compared to the theme arguments. Based on this observation, we can infer that the recognition performance of the "Regu" event is influenced by the recognition performance of the cause arguments.

\begin{table}
\centering
\caption{The recognition performance for different types of trigger words in the dev sets}
\label{trigger_recognition}
\setlength{\tabcolsep}{1.5mm}{
\begin{tabular}{lccc|ccc}
\hline
\multirow{2}{*}{\textbf{Type}} & \multicolumn{3}{c}{\textbf{GE11}} & \multicolumn{3}{c}{\textbf{GE13}} \\
\cline{2-7}
& \textbf{P}(\%) & \textbf{R}(\%) & \textbf{F1}(\%) & \textbf{P}(\%) & R(\%) & \textbf{F1}(\%) \\
\hline
GeEx &83.83 &83.92 &83.84 &78.84 &81.09 &79.94 \\
Tran &70.90 &60.50 &65.28 &61.23 &50.29 &55.17 \\
PrCa &86.06 &98.10 &91.60 &81.91 &72.00 &76.60 \\
Phos &91.86 &91.08 &91.45 &96.32 &94.32 &95.31 \\
Loca &79.95 &86.82 &83.22 &95.53 &83.58 &89.14 \\
Bind &75.67 &74.09 &74.86 &85.22 &81.20 &83.16 \\
Regu &73.14 &67.42 &70.12 &60.46 &59.62 &59.97 \\
PoRe &73.19 &73.09 &73.11 &75.36 &74.65 &74.98 \\
NeRe &80.80 &77.63 &79.17 &83.09 &81.75 &82.39 \\
\hline
Avg. &77.93 &76.38 &77.14 &79.56 &77.58 &78.55 \\
\hline
\end{tabular}
}
\end{table}

\begin{table}
\centering
\caption{The recognition performance for different types of arguments in the dev sets}
\label{argument_recognition}
\setlength{\tabcolsep}{1.5mm}{
\begin{tabular}{lccc|ccc}
\hline
\multirow{2}{*}{\textbf{Type}} & \multicolumn{3}{c}{\textbf{GE11}} & \multicolumn{3}{c}{\textbf{GE13}} \\
\cline{2-7}
& \textbf{P}(\%) & \textbf{R}(\%) & \textbf{F1}(\%) & \textbf{P}(\%) & R(\%) & \textbf{F1}(\%) \\
\hline
Theme &80.94 &74.95 &77.81 &77.01 &70.73 &73.72 \\
Cause &70.14 &53.39 &60.58 &68.06 &45.90 &54.57 \\
\hline
Avg.  &79.73 &72.13 &75.72 &75.83 &66.87 &71.06 \\
\hline
\end{tabular}
}
\vspace{-0.5cm}
\end{table}

\subsubsection{Complexity analysis} To simplify the workflow of biomedical event extraction, we use a data-driven method by introducing the Merging layer. Here, we mathematically prove that the cost of the Merging layer is acceptable from two aspects, time complexity and space complexity.
\begin{itemize}
\item  Time Complexity: We measure time complexity using Floating-point Operations (FLOPs). Without loss of generality, we only consider General Matrix Multiplications (GEMMs) as they are the main component of floating-point operations. Here, we take the self-attention merging method, which has the highest computational complexity, as an example. It consists of 3 steps. First, the total FLOPs of query/key/value matrix transformation (i.e., $Wx$ term) is $2\times(2b\times d_h\times \lvert C\rvert + s\times d_h \times d)$. Second, the total FLOPs of self-attention score (i.e., $QK^\top$ term) is $2\times(b\times\lvert C\rvert \times d_h \times s)$. Third, the total FLOPs of self-attention weighted sum of value matrix (i.e., $AV$ term) is $2\times(b \times s \times \lvert C\rvert \times d_h)$. In detail, $b$ is the batch size, $s$ is the sequence length of an instance with a maximum of 512, $d_h$ is the hidden size of the query/key/value matrix which is set to 768, $d$ is the hidden size of the model which is also set to 768 and $\lvert C\rvert$ is the average size (up to the maximum of 4) of the candidate trigger words of each token. Thus, the FLOPs of the Merging layer is about $768.375M$ FLOPs. When using GPU\footnote{https://developer.nvidia.com/cuda-gpus}, the forward propagation process of the Merging layer consumes much less than 1 second.
\item Space Complexity: We measure space complexity using training parameters. Here, we take the self-attention merging method, which has the most training parameters, as an example. In self-attention merging method, we merely introduce a query matrix $W_Q$, key matrix $W_K$ and a value matrix $W_V$, where $W_Q,W_K,W_V \in \mathbb{R}^{d_h\times d}$. Since $d_h$ and $d$ are set to 768, the training parameters of the self-attention merging method is $3\times d_h\times d = 3\times768^2 = 1769472$, approximately $1.69M$ additional parameters with 1.54\% increase\footnote{The backbone of MLSL is Biobert-base, which contains $110M$ parameters}.
\end{itemize}

\section{Related works}
Biochemical event extraction (i.e., BEE) is a long-standing traditional and important task in the NLP domain~\cite{riedel2011robust,miwa2012boosting,venugopal2014relieving}. Such a task has been dominated by deep learning methods in recent years because neural networks can automatically capture complex features and eliminate the need for feature engineering~\cite{frisoni2021survey}. There are two main deep learning paradigms used to solve biochemical event extraction: pipeline and joint. 

The pipeline method uses an encoder like BiLSTM~\cite{li2016biomedical}, CNN~\cite{bjorne2018biomedical} or a Bert-like model (e.g., SciBERT~\cite{beltagy2019scibert}) to encode the input text. Then it sequentially extracts trigger words, arguments and argument roles, in which errors are accumulated. To better capture the sentence information, dependency parsing tree~\cite{bjorne2018biomedical} and abstract meaning representation~\cite{rao2017biomedical} are introduced to model the semantic or syntactic information. In addition, external knowledge has been proven to be beneficial for improving the performance of BEE~\cite{li2019biomedical}. Through analyzing the data distribution, Wu et al.~\cite{wu2024pipelined} found that the "Binding" type events have a significant impact on the results of the GE11 and GE13 corpora.

The joint method is proposed to address the issue of error accumulation in pipeline methods~\cite{venugopal2014relieving}. Through parameter sharing, it can also reduce computational costs and enhance information exchange between subtasks (i.e., trigger words recognition and arguments recognition)~\cite{trieu2020deepeventmine,zhao2021novel}. However, in order to identify nested events, it is still necessary to use complex structures (e.g.  graphs~\cite{huang2020biomedical,wang2022conditional}) or multi-turn reinforced agent~\cite{zhao2021improved}  to construct the joint method. Ramponi et al.~\cite{ramponi2020biomedical} designed sophisticated label schema to cast joint BEE into a sequence labeling problem. Wang et al.~\cite{wang2020biomedical} transformed BEE into a multi-round question answering task, sequentially identifying the trigger word corresponding to the given entity and other arguments of that predicted trigger word.

\section{Conclusion and future works}
In this work, we propose the MLSL, a system based on multi-layer sequence labeling for joint biomedical event extraction. It is a data-driven method, which does not introduce prior knowledge and complex structures, merging explicitly the information of candidate trigger words into sequence labeling, leading to excellent performance in extensive experiments. However, not all event types and argument types can be effectively identified by the MLSL. Therefore, in the future, we will attempt to use data augmentation methods to address the issue of imbalanced class distribution of data. Based on that, we will also try using models with stronger generalization capabilities to solve BEE tasks.

\section{Acknowledgements}
This work is supported by the National Natural Science Foundation of China [61976147] and the research grants of The Hong Kong Polytechnic
University (\#P0048932, \#P0051089) and A Project Funded by the Priority Academic
Program Development of Jiangsu Higher Education Institutions (PAPD).

\bibliographystyle{splncs04}
% \bibliography{refs.bib}

\end{document}